\documentclass[twoside,11pt]{article}

\usepackage{include/jmlr2e}
\newtheorem{assumption}{Assumption}
\usepackage{microtype}
\usepackage{scalerel}

\usepackage{graphicx}
\usepackage{booktabs} % for professional tables
\usepackage{titletoc}
\usepackage{amsmath}
\usepackage{amssymb}
\usepackage{hyperref}
\usepackage{cleveref}

\usepackage{multirow}
\usepackage{array}
\usepackage{multicol}
\usepackage{tabularx}

\usepackage{hyperref}

\usepackage{amsmath,amsfonts,bm,amssymb}
\def\eqref#1{equation~\ref{#1}}
\def\1{\bm{1}}

\def\rb{{\textnormal{b}}}

\def\vb{{\bm{b}}}

\DeclareMathAlphabet{\mathsfit}{\encodingdefault}{\sfdefault}{m}{sl}
\SetMathAlphabet{\mathsfit}{bold}{\encodingdefault}{\sfdefault}{bx}{n}

\let\ab\allowbreak

\usepackage{latexsym,eucal,bbm,color}
\usepackage{url}
\usepackage{comment}
\usepackage{float}
\usepackage{tcolorbox}
\usepackage{booktabs}
\usepackage{blindtext}
\usepackage{hyperref}
\hypersetup{
    colorlinks,
    linkcolor={red!50!black},
    citecolor={blue!50!black},
    urlcolor={blue!80!black}
}
\usepackage{multirow}
\usepackage{booktabs}
\usepackage{array}
\usepackage{soul}
 
\usepackage[normalem]{ulem}
\usepackage{stackengine}
\usepackage{parskip}
\usepackage{bm}
\usepackage{balance}
\usepackage{comment}
\usepackage{soul}
\usepackage{pgffor}

\usepackage[ruled,vlined]{algorithm2e}

\def \bx{\boldsymbol{x}}

\def \bm{\boldsymbol{m}}

\def \bb{\boldsymbol{b}}

\def \bm{\boldsymbol{m}}

\def \bC{\boldsymbol{C}}
\def \bZ{\boldsymbol{Z}}

\usepackage[textsize=tiny]{todonotes}

\definecolor{mydarkblue}{rgb}{0,0.08,0.45}
\hypersetup{ %
    pdftitle={},
    pdfauthor={},
    pdfsubject={},
    pdfkeywords={},
    pdfborder=0 0 0,
    pdfpagemode=UseNone,
    colorlinks=true,
    linkcolor=mydarkblue,
    citecolor=mydarkblue,
    filecolor=mydarkblue,
    urlcolor=mydarkblue,
    pdfview=FitH
}

\ShortHeadings{To Compress or Not to Compress}{}
\editor{Amos Storkey}

\firstpageno{1}

\begin{document}

\title{To Compress or Not to Compress - Self-Supervised Learning and Information Theory: A Review}

\author{\name Ravid Shwartz-Ziv \addr New York University \email ravid.shwartz.ziv@nyu.edu \\
        \name Yann LeCun \addr New York University \& Meta AI - FAIR \email
        }

\maketitle

%% ABSTRACT
% \begin{abstract}%   <- trailing '%' for backward compatibility of .sty file
\vspace*{-10pt}
\begin{abstract}
Deep neural networks excel in supervised learning tasks but are constrained by the need for extensive labeled data. Self-supervised learning emerges as a promising alternative, allowing models to learn without explicit labels. Information theory, and notably the information bottleneck principle, has been pivotal in shaping deep neural networks. This principle focuses on optimizing the trade-off between compression and preserving relevant information, providing a foundation for efficient network design in supervised contexts. However, its precise role and adaptation in self-supervised learning remain unclear. In this work, we scrutinize various self-supervised learning approaches from an information-theoretic perspective, introducing a unified framework that encapsulates the \textit{self-supervised information-theoretic learning problem}. We weave together existing research into a cohesive narrative, delve into contemporary self-supervised methodologies, and spotlight potential research avenues and inherent challenges. Additionally, we discuss the empirical evaluation of information-theoretic quantities and their estimation methods. Overall, this paper furnishes an exhaustive review of the intersection of information theory, self-supervised learning, and deep neural networks.
\end{abstract}

% \end{abstract}


%%% INTRODUCTION
\section{Introduction}
\label{sec:intro}

Deep neural networks (DNNs) have revolutionized fields such as computer vision, natural language processing, and speech recognition due to their remarkable performance in supervised learning tasks \citep{alam2020survey, natureDeepLeraning, HeZRS15}. However, the success of DNNs is often limited by the need for vast amounts of labeled data, which can be both time-consuming and expensive to acquire. Self-supervised learning (SSL) emerges as a promising alternative, enabling models to learn from data without explicit labels by leveraging the underlying structure and relationships within the data itself.

Recent advances in SSL have been driven by joint embedding architectures, such as Siamese Nets \citep{bromley1993signature}, DrLIM \citep{chopra-cvpr-05,hadsell-cvpr-06}, and SimCLR \citep{chen2020simple}. These approaches define a loss function that encourages representations of different versions of the same image to be similar while pushing representations of distinct images apart. After optimizing the surrogate objective, the pre-trained model can be employed as a feature extractor, with the learned features serving as inputs for downstream supervised tasks like image classification, object detection, instance segmentation, or pose estimation \citep{caron2021emerging, chen2020simple, misra2020self,shwartz2022pre}. Although SSL methods have shown promising results in practice, the theoretical underpinnings behind their effectiveness remain an open question \citep{arora2019theoretical, lee2021predicting}.

Information theory has played a crucial role in understanding and optimizing deep neural networks, from practical applications like the variational information bottleneck \citep{vib} to theoretical investigations of generalization bounds induced by mutual information \citep{xu2017information, steinke2020reasoning}. Building upon these foundations, several researchers have attempted to enhance self-supervised and semi-supervised learning algorithms using information-theoretic principles, such as the Mutual Information Neural Estimator (MINE) \citep{belghazi2018mine} combined with the information maximization (InfoMax) principle \citep{linsker1988self}. However, the plethora of objective functions, contradicting assumptions, and various estimation techniques in the literature can make it challenging to grasp the underlying principles and their implications.

In this paper, we aim to achieve two objectives. First, we propose a unified framework that synthesizes existing research on self-supervised and semi-supervised learning from an information-theoretic standpoint. This framework allows us to present and compare current methods, analyze their assumptions and difficulties, and discuss the optimal representation for neural networks in general and self-supervised networks in particular. Second, we explore different methods and estimators for optimizing information-theoretic quantities in deep neural networks and investigate how recent models optimize various theoretical-information terms.

By reviewing the literature on various aspects of information-theoretic learning, we provide a comprehensive understanding of the interplay between information theory, self-supervised learning, and deep neural networks. We discuss the application of the information bottleneck principle \citep{ib}, connections between information theory and generalization, and recent information-theoretic learning algorithms. Furthermore, we examine how the information-theoretic perspective can offer insights into the design of better self-supervised learning algorithms and the potential benefits of using information theory in SSL across a wide range of applications.

In addition to the main structure of the paper, we dedicate a section to the challenges and opportunities in extending the information-theoretic perspective to other learning paradigms, such as energy-based models. We highlight the potential advantages of incorporating these extensions into self-supervised learning algorithms and discuss the technical and conceptual challenges that must be addressed.

The structure of the paper is as follows. Section 2 introduces the key concepts in supervised, semi-supervised, self-supervised learning, information theory, and representation learning. Section 3 presents a unified framework for multiview learning based on information theory. We first discuss what an optimal representation is and why compression is beneficial for learning. Next, we explore optimal representation in single-view supervised learning models and how they can be extended to unsupervised, semi-supervised, and multiview contexts. The focus then shifts to self-supervised learning, where the optimal representation remains an open question. Using the unified framework, we compare recent self-supervised algorithms and discuss their differences. We analyze the assumptions behind these models, their effects on the learned representation, and their varying perspectives on important information within the network.

Section 5 addresses several technical challenges, discussing both theoretical and practical issues in estimating theoretical information terms. We present recent methods for estimating these quantities, including variational bounds and estimators. Section 6 concludes the paper by offering insights into potential future research directions at the intersection of information theory, self-supervised learning, and deep neural networks. Our aim is to inspire further research that leverages information theory to advance our understanding of self-supervised learning and to develop more efficient and effective models for a broad range of applications.

%% BACKGROUND
\section{Background and Fundamental Concepts}

\subsection{Multiview Representation Learning}
Multiview learning has gained increasing attention and great practical success by using complementary information from multiple features or modalities. The multiview learning paradigm divides the input variable into multiple views from which the target variable should be predicted \citep{ZHAO201743}. Using this paradigm, one can eliminate hypotheses that contradict predictions from other views and provide a natural semi-supervised and self-supervised learning setting. A multiview dataset consists of data captured from multiple sources, modalities, and forms but with similar high-level semantics \citep{YAN2021106}. This mechanism was initially used for natural-world data, combining image, text, audio, and video measurements. For example, photos of objects are taken from various angles, and our supervised task is to identify the objects. Another example is identifying a person by analyzing the video stream as one view and the audio stream as the other.

Although these views often provide different and complementary information about the same data, directly integrating them does not produce satisfactory results due to biases between multiple views \citep{YAN2021106}. Thus, multiview representation learning involves identifying the underlying data structure and integrating the different views into a common feature space, resulting in high performance. In recent decades, multiview learning has been used for many machine learning tasks and influenced many algorithms, such as co-training mechanisms \citep{kumar2011co}, subspace learning methods \citep{XUE2019210}, and multiple kernel learning (MKL) \citep{bach2002kernel}. \cite{li2018survey} proposed two categories for multiview representation learning: (i) multiview representation fusion, which combines different features from multiple views into a single compact representation, and (ii) alignment of multiview representation, which attempts to capture the relationships among multiple different views through feature alignment. In this case, a learned mapping function embeds the data of each view, and the representations are regularized to form a multiview-aligned space. In this research direction, an early study is the Canonical Correlation Analysis (CCA) \citep{cca1396} and its kernel extensions \citep{kernel2003,hardoon2004, Sun2013ASO}. In addition to CCA, multiview representation learning has penetrated a variety of learning methods, such as dimensionality reduction \citep{sun2010scalable}, clustering analysis \citep{yan2015unsupervised}, multiview sparse coding \citep{factorized2010, Tian2013, multi2014}, and multimodal topic learning \citep{pu2020multimodal}. However, despite their promising results, these methods use handcrafted features and linear embedding functions, which cannot capture the nonlinear properties of multiview data.

The emergence of deep learning has provided a powerful way to learn complex, nonlinear, and hierarchical representations of data. By incorporating multiple hierarchical layers, deep learning algorithms can learn complex, subtle, and abstract representations of target data. The success of deep learning in various application domains has led to a growing interest in deep multiview methods, which have shown promising results. Examples of these methods include deep multiview canonical correlation analysis \citep{andrew2013deep} as an extension of CCA, multiview clustering via deep matrix factorization \citep{zhao2017multi}, and the deep multiview spectral network \citep{huang2019multi}. Moreover, deep architectures have been employed to generate effective representations in methods such as multiview convolutional neural networks \citep{liu2021deep}, multimodal deep Boltzmann machines \citep{srivastava14b}, multimodal deep autoencoders \citep{multimodel2011, deepmultiview2015}, and multimodal recurrent neural networks \citep{karpathy2015deep, mao2014deep, donahue2015long}.

\subsection{Self-Supervised Learning}

Self-supervised learning (SSL) is a powerful technique that leverages unlabeled data to learn useful representations. In contrast to supervised learning, which relies on labeled data, SSL employs self-defined signals to establish a proxy objective between the input and the signal. The model is initially trained using this proxy objective and subsequently fine-tuned on the target task. Self-supervised signals, derived from the inherent co-occurrence relationships in the data, serve as self-supervision. Various such signals have been used to learn representations, including generative and joint embedding architectures \citep{chen2020simple, chen2020improved, bachman2019learning, bar2022detreg}.

Two main categories of SSL architectures exist: (1) generative architectures based on reconstruction or prediction and (2) joint embedding architectures \citep{liu2021self}. Both architecture classes can be trained using either contrastive or non-contrastive methods.

We begin by discussing these two main types of architectures:

\begin{enumerate}

\item \textbf{Generative Architecture:} Generative architectures employ an objective function that measures the divergence between input data and predicted reconstructions, such as squared error. The architecture reconstructs data from a latent variable or a corrupted version, potentially with a latent variable's assistance. Notable examples of generative architectures include auto-encoders, sparse coding, sparse auto-encoders, and variational auto-encoders~\citep{kingma2013auto, NIPS2006_2d71b2ae, ng2011sparse}. As the reconstruction task lacks a single correct answer, most generative architectures utilize a latent variable, which, when varied, generates multiple reconstructions. The latent variable's information content requires regularization to ensure the system reconstructs regions of high data density while avoiding a collapse by reconstructing the entire space. PCA regularizes the latent variable by limiting its dimensions, while sparse coding and sparse auto-encoders restrict the number of non-zero components. Variational auto-encoders regularize the latent variable by rendering it stochastic and maximizing the entropy of the distribution relative to a prior. Vector quantized variational auto-encoders (VQ-VAE) employ binary stochastic variables to achieve similar results~\citep{oord2017neural}.

\item \textbf{Joint Embedding Architectures (JEA):} These architectures process multiple views of an input signal through encoders, producing representations of the views. The system is trained to ensure that these representations are both informative and mutually predictable. Examples include Siamese networks, where two identical encoders share weights~\citep{chen2021exploring,chen2020simple,he2020momentum,grill2020bootstrap}, and methods permitting encoders to differ~\citep{bardes2021vicreg}. A primary challenge with JEA is preventing informational collapse, in which the representations contain minimal information about the inputs, thereby facilitating their mutual prediction. JEA's advantage lies in the encoders' ability to eliminate noisy, unpredictable, or irrelevant information from the input within the representation space.

\end{enumerate}

To effectively train these architectures, it is essential to ensure that the representations of different signals are distinct. This can be achieved through either contrastive or non-contrastive methods:

\begin{itemize}
\item \textbf{Contrastive Methods:} Contrastive methods utilize data points from the training set as \textit{positive samples} and generate points outside the region of high data density as \textit{contrastive samples}. The energy (e.g., reconstruction error for generative architectures or representation predictive error for JEA) should be low for positive samples and higher for contrastive samples. Various loss functions involving the energies of pairs or sets of samples can be minimized to achieve this objective.

\item \textbf{Non-Contrastive Methods:} Non-contrastive methods prevent the energy landscape's collapse by limiting the volume of space that can take low energy, either through architectural constraints or through a regularizer in the energy or training objective. In latent-variable generative architectures, preventing collapse is achieved by limiting or minimizing the information content of the latent variable. In JEA, collapse is prevented by maximizing the information content of the representations.
\end{itemize}

We now present a few concrete examples of popular models that employ various combinations of generative architectures, joint embedding architectures, contrastive training, and non-contrastive training:

The \textbf{Denoising Autoencoder} approach in generative architectures~\citep{vincent2008extracting,devlin2018bert,he2022masked} using a triplet loss which utilizes a positive sample, which is a vector from the training set that should be reconstructed perfectly, and a contrastive sample consisting of data vectors, one from the training set and the other being a corrupted version of it.  In SSL, the combination of \textit{JEA} models with \textit{contrastive learning} has proven highly effective. In contrastive learning, the objective is to attract different augmented views of the same image (positive points) while repelling dissimilar augmented views (negative points). Recent self-supervised visual representation learning examples include MoCo \citep{he2020momentum} and SimCLR \citep{chen2020simple}. The InfoNCE loss is a commonly used objective function in many contrastive learning methods:
\begin{align*}
\mathbb{E}_{x,x^+, x^-}\left[-\log \left(\frac{e^{f(x)^Tf(x^+)}}{\sum{k=1}^K{e^{f(x)^Tf(x^k)}}}\right)\right]
\end{align*}
where $x+$ is a sample similar to $x$, $x^k$ are all the samples in the batch, and $f$ is an encoder.

However, contrastive methods heavily depend on all other samples in the batch and require a large batch size. Additionally, recent studies \citep{jing2021understanding} have shown that contrastive learning can lead to dimensional collapse, where the embedding vectors span a lower-dimensional subspace instead of the entire embedding space. Although positive and negative pairs should repel each other to prevent dimensional collapse, augmentation along feature dimensions and implicit regularization cause the embedding vectors to fall into a lower-dimensional subspace, resulting in low-rank solutions.

To address these problems, recent works have introduced \textit{JEA} models with \textit{non-contrastive methods}. Unlike contrastive methods, these methods employ regularization to prevent the collapse of the representation and do not explicitly rely on negative samples. For example, several papers use stop-gradients and extra predictors to avoid collapse \citep{chen2021exploring, grill2020bootstrap}, while \cite{caron2020unsupervised} employed an additional clustering step. VICReg \citep{bardes2021vicreg} is another non-contrastive method that regularizes the covariance matrix of representation. Consider two embedding batches $\bZ=\left[f(\bx_1),\dots,f(\bx_N)\right]$ and $\bZ^\prime=\left[f(\bx'1),\dots,f(\bx'N) \right]$, each of size $(N \times K)$. Denote by $\bC$ the $(K \times K)$ covariance matrix obtained from $[\bZ,\bZ']$. The VICReg triplet loss is defined by:
\begin{gather*}
    \mathcal{L}\hspace{-0.1cm}=\frac{1}{K}\sum_{k=1}^K\hspace{-0.1cm}\left(\hspace{-0.1cm}\alpha\max \left(0, \gamma- \sqrt{\bC_{k,k} +\epsilon}\right)\hspace{-0.1cm}+\hspace{-0.1cm}\beta \sum_{k'\neq k}\hspace{-0.1cm}\left(\bC_{k,k'}\right)^2\hspace{-0.1cm}\right)
    \;\;\;+\gamma\| \bZ-\bZ'\|_F^2/N.
\end{gather*}

\subsection{Semi-Supervised Learning}
Semi-supervised learning employs both labeled and unlabeled data to enhance the model performance \citep{chapelle2009semi}. Consistency regularization-based approaches \citep{laine2016temporal,miyato2018virtual, sohn2020fixmatch} ensure that predictions remain stable under perturbations in input data and model parameters. Certain techniques, such as those proposed by \cite{grandvalet2006entropy} and \cite{miyato2018virtual}, involve training a model by incorporating a regularization term into a supervised cross-entropy loss. In contrast, \cite{xie2020unsupervised} utilizes suitably weighted unsupervised regularization terms, while \cite{zhai2019s4l} adopts a combination of self-supervised pretext loss terms. Moreover, pseudo-labeling can generate synthetic labels based on network uncertainty to further aid model training \citep{Lee2013PseudoLabelT}.

%% representation learning
\subsection{Representation Learning}
\label{sec:learning}
Representation learning is an essential aspect of various computer vision, natural language processing, and machine learning tasks, as it uncovers the underlying structures in data \citep{bengio2013representation}. By extracting relevant information for classification and prediction tasks from the data, we can improve performance and reduce computational complexity \citep{Goodfellow-et-al-2016-Book}. However, defining an effective representation remains a challenging task. In probabilistic models, a useful representation often captures the posterior distribution of explanatory factors beneath the observed input \citep{natureDeepLeraning}. \cite{BengioAndLecunMIT} introduced the idea of learning highly structured yet complex dependencies for AI tasks, which require transforming high-dimensional input structures into low-dimensional output structures or learning low-level representations. As a result, identifying relevant input features becomes challenging, as most input entropy is unrelated to the output \citep{shwartz2017opening}. \cite{ben2023reverse} demonstrated that self-supervised learning inherently promotes the clustering of samples based on semantic labels. Intriguingly, this clustering is driven by the objective's regularization term and aligns with semantic classes across multiple hierarchical levels.

\subsubsection{Minimal Sufficient Statistic}

A possible definition of an effective representation is based on  \textit{minimal sufficient statistics.}

\begin{definition}
Given $(X,Y)\sim P(X,Y)$, let $T:=t(X)$, where $t$ is a deterministic function. We define $T$ as a sufficient statistic of $X$ for $Y$ if $Y-T-X$ forms a Markov chain.
\end{definition}

A sufficient statistic captures all the information about $Y$ in $X$. \cite{cover1999elements} proved this property:

\begin{theorem}
Let $T$ be a probabilistic function of $X$. Then, $T$ is a sufficient statistic for $Y$ if and only if $I(T(X); Y )=I(X; Y )$.
\end{theorem}

However, the sufficiency definition also encompasses trivial identity statistics that only "copy" rather than "extract" essential information. To prevent statistics from inefficiently utilizing observations, the concept of minimal sufficient statistics was introduced:

\begin{definition}(Minimal sufficient statistic (MSS))
A sufficient statistic $T$ is minimal if, for any other sufficient statistic $S$, there exists a function $f$ such that $T=f(S)$ almost surely (a.s.).
\end{definition}

In essence, MSS are the simplest sufficient statistics, inducing the coarsest sufficient partition on $X$. In MSS, the values of $X$ are grouped into as few partitions as possible without sacrificing information. MSS are statistics with the maximum information about $Y$ while retaining the least information about $X$ as possible \citep{koopman1936distributions}.

\subsubsection{The Information Bottleneck}
\label{sec:ib}
The majority of distributions lack exact minimal sufficient statistics, leading \cite{Tishby1999} to relax the optimization problem in two ways: (i) allowing the map to be stochastic, defined as an encoder $P(T| X)$, and (ii) permitting the capture of only a small amount of $I(X; Y)$. The information bottleneck (IB) was introduced as a principled method to extract relevant information from observed signals related to a target. This framework finds the optimal trade-off between the accuracy and complexity of a random variable $y \in \mathcal{Y}$ with a joint distribution for a random variable $x\in \mathcal{X}$. The IB has been employed in various fields such as neuroscience \citep{buesing2010spiking,palmer2015predictive}, slow feature analysis \citep{turner2007maximum}, speech recognition \citep{hecht2009speaker}, and deep learning \citep{shwartz2017opening, vib}.

Let $X$ be an input random variable, $Y$ a target variable, and $P(X,Y)$ their joint distribution. A representation $T$ is a stochastic function of $X$ defined by a mapping $P(T \mid X)$. This mapping transforms $X \sim P(X)$ into a representation of $T\sim P(T):=\int P_{T \mid X}(\cdot \mid x)dP_X(x)$. The triple $Y-X-T$ forms a Markov chain in that order with respect to the joint probability measure $P_{X,Y,T}=P_{X,Y}P_{T \mid X}$ and the mutual information terms $I(X; T)$ and $I(Y; T)$.

Within the IB framework, our goal is to find a representation $P(T \mid X)$ that extracts as much information as possible about $Y$ (high performance) while compressing $X$ maximally (keeping $I(X; T)$ small). This can also be interpreted as extracting only the relevant information that $X$ contains about $Y$.

The data processing inequality (DPI) implies that $I(Y; T)\leq I(X; Y)$, so the compressed representation $T$ cannot convey more information than the original signal. Consequently, there is a trade-off between compressed representation and the preservation of relevant information about $Y$. The construction of an efficient representation variable is characterized by its encoder and decoder distributions, $P(T \mid X)$ and $P(Y \mid T)$, respectively. The efficient representation of $X$ involves minimizing the complexity of the representation $I\left(T; X\right)$ while maximizing $I\left(T; Y\right)$. Formally, the IB optimization involves minimizing the following objective function:
\begin{align}
\label{eq:IB}
\mathcal{L}=\min_{P(t \mid x); p(y \mid t )} I(X;T) - \beta I(Y;T)~,
\end{align}
where $\beta$ is the trade-off parameter controlling the complexity of $T$ and the amount of relevant information it preserves. Intuitively, we pass the information that $X$ contains about $Y$ through a ``bottleneck'' via the representation $T$.
It has been shown that:
\begin{align}
I(T:Y)=I(X:Y)-\mathbb{E}_{x\sim P(X), t\sim P(T|x)}\left[D\left[P(Y|x)||P(Y|t)\right]\right]
\end{align}

\subsection{Representation Learning and the Information Bottleneck}
\label{sec:learning_bi}

Information theory traditionally assumes that underlying probabilities are known and do not require learning. For instance, the optimality of the initial IB work \citep{Tishby1999} relied on the assumption that the joint distribution of input and labels is known. However, a significant challenge in machine learning algorithms is inferring an accurate predictor for the unknown target variable from observed realizations. This discrepancy raises questions about the practical optimality of the IB and its relevance in modern learning algorithms. The following section delves into the relationship between the IB framework and learning, inference, and generalization.

Let $X\in \mathcal{X}$ and a target variable $Y\in\mathcal{Y}$ be random variables with an unknown joint distribution $P(X, Y)$. For a given class of predictors $f:\mathcal{X}\to \mathcal{\hat{Y}}$ and a loss function $\ell:\mathcal{Y} \to \mathcal{\hat{Y}}$ measuring discrepancies between true values and model predictions, our objective is to find the predictor $f$ that minimizes the expected population risk.
\begin{align*}
\mathcal{L}_{P(X,Y)}\left(f, \ell\right) = \mathbb{E}_{P(X,Y)}\left[\ell(Y, f(X))\right]
\end{align*}

Several issues arise with the population risk. Firstly, it remains unclear which loss function is optimal. A popular choice is the logarithmic loss (or error's entropy), which has been numerically demonstrated to yield better results \citep{erdogmus2002information}. This loss has been employed in various algorithms, including the InfoMax principle \citep{linsker1988self}, tree-based algorithms \citep{quinlan2014c4}, deep neural networks \citep{zhang2018generalized}, and Bayesian modeling \citep{wenzel2020good}. \cite{painsky2018universality} provided a rigorous justification for using the logarithmic loss and showed that it is an upper bound to any choice of the loss function that is smooth, proper, and convex for binary classification problems.

In most cases, the joint distribution $P(X, Y)$ is unknown, and we have access to only $n$ samples from it, denoted by $\mathcal{D}_n := {(x_i, y_i) \mid i=1,\ldots,n}$.  Consequently, the population risk cannot be computed directly. Instead, we typically choose the predictor that minimizes the empirical population risk on a training dataset:

\begin{align*}
\mathcal{\hat{L}}_{P(X,Y)}\left(f, \ell, \mathcal{D}_n\right) = \frac{1}{n}\sum_{i=1}^n\left[\ell(y_i, f(x_i))\right]
\end{align*}

The generalization gap, defined as the difference between empirical and population risks, is given by: 
\begin{align*}
Gen_{P(X,Y)}\left(f, \ell, \mathcal{D}_n\right) := \mathcal{L}_{P(X,Y)}\left(f, \ell\right) - \mathcal{\hat{L}}_{P(X,Y)}\left(f, \ell, \mathcal{D}_n\right)
\end{align*}

Interestingly, the relationship between the true loss and the empirical loss can be bounded using the information bottleneck term. \cite{SHAMIR20102696} developed several finite sample bounds for the generalization gap. According to their study, the IB framework exhibited good generalizability even with small sample sizes. In particular, they developed non-uniform bounds adaptive to the model's complexity. They demonstrated that for the discrete case, the error in estimating mutual information from finite samples is bounded by $O\left(\frac{|X|\log n}{\sqrt{n}}\right)$, where $|X|$ is the cardinality of $X$ (the number of possible values that the random variable $X$ can take). The results support the intuition that simpler models generalize better, and we would like to compress our model. Therefore, optimizing \cref{eq:IB} presents a trade-off between two opposing forces. On the one hand, we want to increase our prediction accuracy in our training data (high $\beta$).

On the other hand, we would like to decrease $\beta$ to narrow the generalization gap. \cite{vera2018role} extended their work and showed that the generalization gap is bounded by the square root of mutual information between training input and model representation times $\frac{\log n}{n}$. Furthermore, \cite{russo2019much} and \cite{xu2017information} demonstrated that the square root of the mutual information between the training input and the parameters inferred from the training algorithm provides a concise bound on the generalization gap. However, these bounds critically depend on the Markov operator that maps the training set to the network parameters, whose characterization is not trivial.

\cite{achille2018emergence} explored how applying the IB objective to the network's parameters may reduce overfitting while maintaining invariant representations. Their work showed that flat minima, which have better generalization properties, bound the information with the weights, and the information in the weights bound the information in the activations. \cite{chelombiev2019adaptive} found that the generalization precision is positively correlated with the degree of compression of the last layer in the network. \cite{shwartz2018representation} showed that the generalization error depends exponentially on the mutual information between the model and the input once it is smaller than $\log2n$ - the query sample complexity. Moreover, they demonstrated that $M$ bits of compression of $X$ are equivalent to an exponential factor of $2^M$ training examples. \cite{piran2020dual} extended the original IB to the dual form, which offers several advantages in terms of compression.

These studies illustrate that the IB leads to a trade-off between prediction and complexity, even for the empirical distribution. With the IB objective, we can design estimators to find optimal solutions for different regimes with varying performance, complexity, and generalization.

%% 4_information Theoretic Objectives
\section{Information-Theoretic Objectives}
Before delving into the details, this section aims to provide an overview of the information-theoretic objectives in various learning scenarios, including supervised, unsupervised, and self-supervised settings. We will also introduce a general framework to understand better the process of learning optimal representations and explore recent methods working towards this goal.

Developing a novel algorithm entails numerous aspects, such as architecture, initialization parameters, learning algorithms, and pre-processing techniques. A crucial element, however, is the objective function. As demonstrated in Section \ref{sec:ib}, the IB approach, originally introduced by \citet{Tishby1999}, defines the optimal representation in supervised scenarios, enabling us to identify which terms to compress during learning. However, determining the optimal representation and deriving information-based objective functions in self-supervised settings are more challenging. In this section, we introduce a general framework to understand the process of learning optimal representations and explore recent methods striving to achieve this goal.

\subsection{Setup and Methodology}

Using a two-channel input allows us to model complex multiview learning problems. In many real-world situations, data can be observed from multiple perspectives or modalities, making it essential to develop learning algorithms capable of handling such multiview data.

Consider a two-channel input, $X_1$ and $X_2$, and a single-channel label $Y$ for a downstream task, all possessing a joint distribution $P(X_1,X_2, Y)$. We assume the availability of $n$ labeled examples $S={(x^i_1,x^i_2,y^i)}^n_{i=1}$ and $t$ unlabeled examples $U={(x^i_1, x^i_2)}^{n+t}_{i=n+1}$, both independently and identically distributed. Our objective is to predict $Y$ using a loss function.

In our model, we use a learned encoder with a prior $P(Z)$ to generate a conditional representation (which may be deterministic or stochastic) $Z_i|X_i = P_{\theta_i}(Z_i|X_i)$, where $i=1,2$ represents the two views. Subsequently, we utilize various decoders to 'decode' distinct aspects of the representation:

For the supervised scenario, we have a joint embedding of the label classifiers from both views, $\hat{Y}_{1,2} = Q_{\rho}(Y|Z_1,Z_2)$, and two decoders predicting the labels of the downstream task based on each individual view, $\hat{Y_i} = Q_{\rho_i}(Y|Z_i)$ for $i=1,2$.

For the unsupervised case, we have direct decoders for input reconstruction from the representation, $\bar{X_i} = Q_{\psi_i}(X_i|Z_i)$ for $i=1,2$.

\begin{figure}
\centering
\includegraphics[scale=0.24]{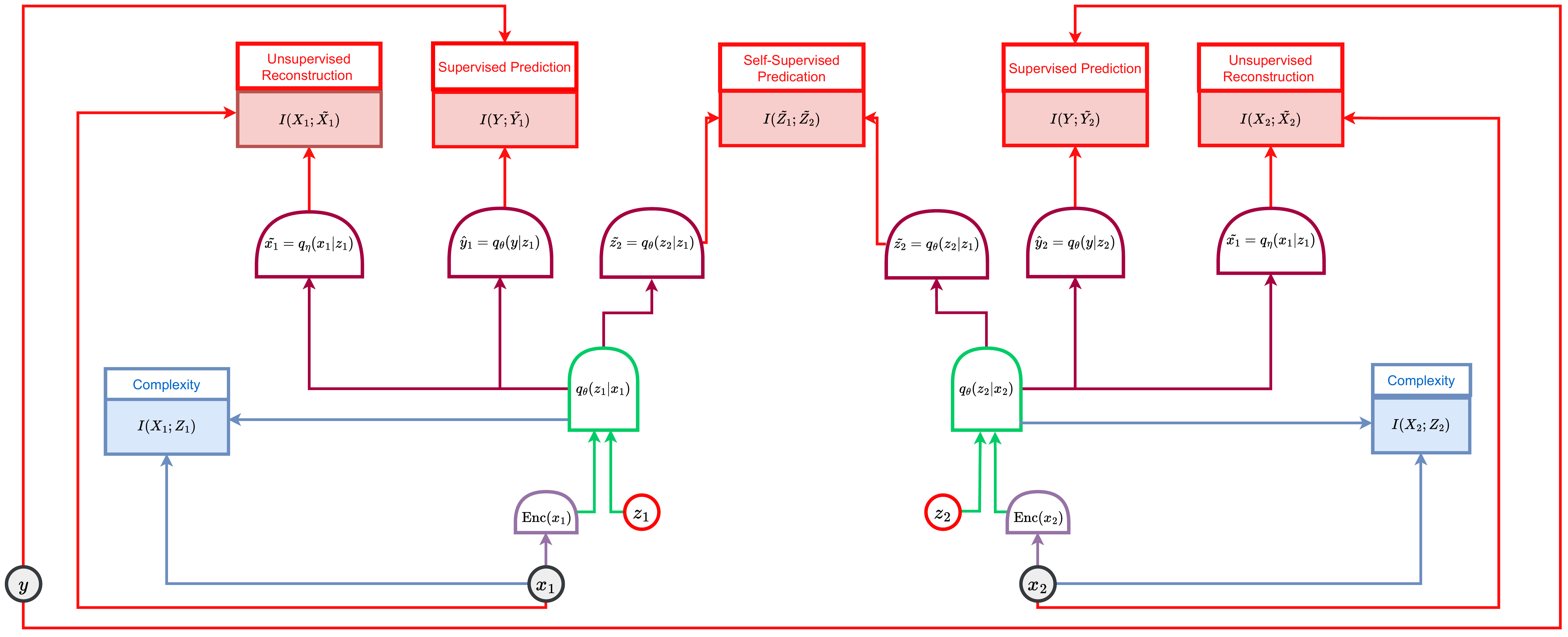}
\caption{Multiview information bottleneck diagram for self-supervised, unsupervised, and supervised learning }
\label{fig:diagram}
\end{figure}

For self-supervised learning, we utilize two cross-decoders attempting to predict one representation based on the other, $\tilde{Z_1}|Z_2=q_{\eta_1}(Z_1|Z_2)$ and $\tilde{Z_2}|Z_1=q_{\eta_2}(Z_2|Z_1)$. Figure \ref{fig:diagram} illustrates this structure.

The information-theoretic perspective of self-supervised networks has led to confusion regarding the information being optimized in recent work. In supervised and unsupervised learning, only one 'information path' exists when optimizing information-theoretic terms: the input is encoded through the network, and then the representation is decoded and compared to the targets. As a result, the representation and corresponding information always stem from a single encoder and decoder.

However, in the self-supervised multiview scenario, we can construct our representation using various encoders and decoders. For instance, we need to specify the associated random variable to define the information involved in $I(X_1; Z_1)$. This variable could either be based on the encoder of $X_1$ - $P_{\theta_1}(Z_1|X_1)$, or based on the encoder of $X_2$ - $P_{\theta_2}(Z_2|X_2)$, which is subsequently passed to the cross-decoder $Q_{\eta_1}(Z_1|Z_2)$ and then to the direct decoder $Q_{\psi_1}(X_1|Z_1)$.

To fully understand the information terms, we aim to optimize and distinguish between various "information paths," we marked each information path differently. For example, $I_{, P(X_1),P(Z_1|X_1),P(Z_2|Z_1)}\left(X_1,Z_2\right)$ is based on the path $P(X_1) \to P(Z_1|X_1) \to P(Z_2|Z_1)$. In the following section, we will "translate" previous work into our present framework and examine the loss function.

\subsection{Optimization with Labels}
After establishing our framework, we can now incorporate various learning algorithms. We begin by examining classical single-view supervised information bottleneck algorithms for deep networks that utilize labeled data during training and extend them to the multiview scenario. Next, we broaden our perspective to include unsupervised learning, where input reconstruction replaces labels, and semi-supervised learning, where information-based regularization is applied to improve predictions.

\subsubsection{Single-View Supervised Learning}
In classical single-view supervised learning, the task of representation learning involves finding a distribution $p(z|x)$ that maps data observations $x \in \mathcal{X}$ to a representation $z \in \mathcal{Z}$, capturing only the relevant features of the input \cite{shwartz2022information}. The goal is to predict a label $y \in \mathcal{Y}$ using the learned representation. \cite{achille2018emergence} defined the sufficiency of $Z$ for $Y$ as the amount of label information retained after passing data through the encoder:

\begin{definition}
\textbf{Sufficiency}: A representation $Z$ of $X$ is sufficient for $Y$ if and only if $I(X;Y|Z)=0$.
\end{definition}

\cite{federici2020learning} showed that $Z$ is sufficient for $Y$ if and only if the amount of information regarding the task remains unchanged by the encoding procedure. A sufficient representation can predict $Y$ as accurately as the original data $X$. In Section \ref{sec:learning}, we saw a trade-off between prediction and generalization when there is a finite amount of data. To reduce the generalization gap, we aim to compress $X$ while retaining as much predicate information on the labels as possible. Thus, we relax the sufficiency definition and minimize the following objective:

\begin{align}
\label{eq:supervisedinf}
\mathcal{L}= I(X;Z) - \beta I(Z;Y)
\end{align}

The mutual information $I(Y;Z)$ determines how much label information is accessible and reflects the model's ability to predict performance on the target task. $I(X;Z)$ represents the information that $Z$ carries about the input, which we aim to compress. However, $I(X;Z)$ contains both relevant and irrelevant information about $Y$. Therefore, using the chain rule of information, \cite{federici2020learning} proposed splitting $I(X,Z)$ into two terms:
\begin{align}
\label{eq:factorize}
I(X;Z) = \underbrace{I(X;Z|Y)}_\text{superfluous information} + \underbrace{I(Z;Y)}_{\text{predictive information}}
\end{align}
The conditional information $I(X,Z|Y)$ represents information in $Z$ that is not predictive of $Y$, i.e., superfluous information. The decomposition of input information enables us to compress only irrelevant information while preserving the relevant information for predicting $Y$. Several methods are available for evaluating and estimating these information-theoretic terms in the supervised case (see Section \ref{sec:estimators} for details).

\subsubsection{The Information Bottleneck Theory of Deep Learning}

The IB hypothesis for deep learning proposes two distinct phases of training neural networks \citep{shwartz2017opening}: the fitting and compression phases. The fitting phase involves extracting information from the input and converting it into learned representations, characterized by increased mutual information between inputs and hidden representations. Conversely, the compression phase, which is much longer, concentrates on discarding unnecessary information for target prediction, decreasing mutual information between learned representations and inputs. In contrast, the mutual information between representations and targets increases. For more information, see \cite{geiger2020information}. Despite the elegance and plausibility of the IB hypothesis, empirically investigating it remains challenging \citep{amjad2018not}.

The study of representation compression in Deep Neural Networks (DNNs) for supervised learning has shown inconsistent results. For instance, \cite{chelombiev2019adaptive} discovered a positive correlation between generalization accuracy and the compression level of the network's final layer. \cite{shwartz2018representation} also examined the relationship between generalization and compression, demonstrating that generalization error exponentially depends on mutual information, $I(X;Z)$. Furthermore, \cite{achille2017critical} established that flat minima, known for their improved generalization properties, constrain the mutual information. However, \cite{saxe2018information} showed that compression was not necessary for generalization in deep linear networks. \cite{Basirat2021} revealed that the decrease in mutual information is essentially equivalent to geometrical compression. Other studies have found that the mutual information between training inputs and inferred parameters provides a concise bound on the generalization gap \citep{xu2017information,pensia2018generalization}. Lastly, \cite{achille2018emergence} explored using an information bottleneck objective on network parameters to prevent overfitting and promote invariant representations.

\subsubsection{Multiview IB Learning}
The IB principle offers a rigorous method for learning encoders and decoders in supervised single-view problems. However, it is not directly applicable to multiview learning problems, as it assumes only one information source as the input. A common solution is to concatenate multiple views, though this neglects the unique characteristics of each view. To address this issue, \cite{Xu2014LargeMarginMB} introduced the large-margin multiview IB (LMIB) as an extension of the original IB problem. LMIB employs a communication system where multiple senders represent various views of examples. The system extracts specific components from different senders by compressing examples through a "bottleneck," and the linear projectors for each view are combined to create a shared representation. The large-margin principle replaces the maximization of mutual information in prediction, emphasizing the separation of samples from different classes. Limiting Rademacher complexity improves the solution's accuracy and generalization error bounds. Moreover, the algorithm's robustness is enhanced when accurate views counterbalance noisy views.

However, the LMIB method has a significant limitation: it utilizes linear projections for each view, which can restrict the combined representation when the relationship between different views is complex. To overcome this limitation, \cite{DeepMultiview2019} proposed using deep neural networks to replace linear projectors. Their model first extracts concise latent representations from each view using deep networks and then learns the joint representation of all views using neural networks. They minimize the objective:
\begin{align*}
\mathcal{L} = \alpha I_{\scaleto{{P(X_1),P(Z_1|X_1})}{6pt}}(X_1;Z_1) + \beta I_{\scaleto{{P(X_2),P(Z_2|X_2})}{6pt}}(X_2;Z_2) - I_{\scaleto{{P(Z_2|X_2),P(Z_2|X_1})}{6pt}}(Z_{1,2};Y)
\end{align*}
Here, $\alpha$ and $\beta$ are trade-off parameters, $Z_1$ and $Z_2$ are the two neural networks' representations, and $Z_{1,2}$ is the joint embedding of $Z_1$ and $Z_2$. The first two terms decrease the mutual information between a view's latent representation and its original data representation, resulting in a simpler and more generalizable model. The final term forces the joint representation to maximize the discrimination ability for the downstream task.

\subsubsection{Semi-Supervised IB Learning: Leveraging Unlabeled Data}
Obtaining labeled data can be challenging or expensive in many practical scenarios, while many unlabeled samples may be readily available. Semi-supervised learning addresses this issue by leveraging the vast amount of unlabeled data during training in conjunction with a small set of labeled samples. Common strategies to achieve this involve adding regularization terms or adopting mechanisms that promote better generalization. \cite{berthelot2019mixmatch} grouped regularization methods into three primary categories: entropy minimization, consistency regularization, and generic regularization.

\cite{voloshynovskiy2019information} introduced an information-theoretic framework for semi-supervised learning based on the IB principle. In this context, the semi-supervised classification problem involves encoding input $X$ into the latent space $Z$ while preserving only \textbf{class-relevant information}. A supervised classifier can achieve this if there is sufficient labeled data. However, when the number of labeled examples is limited, the standard label classifier $p(y|z)$ becomes unreliable and requires regularization.

To tackle this issue, the authors assumed a prior on the class label distribution $p(y)$. They introduced a term to minimize the $D_{KL}$ between the assumed marginal prior and the empirical marginal prior, effectively regularizing the conditional label classifier with the labels' marginal distribution. This approach reduces the classifier's sensitivity to the scarcity of labeled examples. They proposed two  variational IB semi-supervised extensions for the priors:

\textbf{Handcrafted Priors}: These priors are predefined for regularization and can be based on domain knowledge or statistical properties of the data. Alternatively, they can be learned using other networks. Handcrafted priors in this context are similar to priors used in the Variational Information Bottleneck (VIB) formalism \citep{vib, DeepMultiview2019}.

\textbf{Learnable Priors}: \cite{voloshynovskiy2019information} also suggests using learnable priors as an alternative to handcrafted regularization priors on the latent representation. This method involves regularizing $Z$ through another IB-based regularization with two components: (i) latent space regularization and (ii) observation space regularization. In this case, an additional hidden variable $M$ is introduced after the representation to regulate the information flow between $Z$ and $Y$. An auto-encoder $q(m|z)$ is employed, and the optimization process aims to compress the information flowing from $Z$ to $M$ while retaining only label-relevant information. The IB objective is defined as:
\begin{equation}
\begin{aligned}
\label{eq:learnableprior_combined}
\mathcal{L} &= D_{KL}(q(m|z) || p(m|z)) - \beta D_{KL}(q(x|m) || p(x|m)) - \beta_y D_{KL}(p(y|z) || p(y)) \\
&\Leftrightarrow I(M; Z) - \beta I(M; X) - \beta_y I(Y; Z)
\end{aligned}
\end{equation}

Here, $\beta$ and $\beta_y$ are hyperparameters that balance the trade-off between the relevance of $M$ to the labels and the compression of $Z$ into $M$.

Furthermore, \cite{voloshynovskiy2019information} demonstrated that various popular semi-supervised methods can be considered special cases of the optimization problem described above. Notably, the semi-supervised AAE \citep{makhzani2015adversarial}, CatGAN \citep{springenberg2015unsupervised}, SeGMA \citep{smieja1906segma}, and VAE \citep{kingma2014semi} can all be viewed as specific instantiations of this framework.

\subsubsection{Unsupervised IB learning}
In the unsupervised setting, data samples are not directly labeled by classes. \cite{voloshynovskiy2019information} defined unsupervised IB as a 'compressed' parameterized mapping of $X$ to $Z$, which preserves some information in $Z$ about $X$ through the reverse decoder $\bar{X}=Q(X|Z)$. Therefore, the Lagrangian of unsupervised IB can be defined as follows:
\begin{align*}
I_{\scaleto{{P(X),P(Z|X})}{6pt}}(X;Z) - \beta I_{\scaleto{{P(Z),Q(X|Z})}{6pt}}(Z;\bar{X})
\end{align*}
where $I(X;Z)$ is the information determined by the encoder $q(z|x)$ and $I(Z;\bar{X})$ is the information determined by the decoder $q(x|z)$, i.e., the reconstruction error. In other words, unsupervised IB is a special case of supervised IB, where labels are replaced with the reconstruction performance of the training input. \cite{vib} showed that Variational Autoencoder (VAE)  \citep{kingma2019introduction} and $\beta$-VAE \citep{higgins2016beta} are special cases of unsupervised variational IB. \cite{voloshynovskiy2019information} extended their results and showed that many models, including adversarial autoencoders \citep{makhzani2015adversarial}, InfoVAEs \citep{zhao2017infovae}, and VAE/GANs \citep{larsen2016autoencoding}, could be viewed as special cases of unsupervised IB. The main difference between them is the bounds on the different mutual information of the IB. Furthermore, unsupervised IB was used by \cite{uugur2020variational} to derive lower bounds for their unsupervised generative clustering framework, while \cite{roy2018theory} used it to study vector-quantized autoencoders.

\cite{voloshynovskiy2019information} pointed out that for the classification task in supervised IB, the latent space $Z$ should be sufficient statistics for $Y$, whose entropy is much lower than $X$. This results in a highly compressed representation where sequences close in the input space might be close in the latent space, and the less significant features will be compressed. In contrast, in the unsupervised setup, the IB suggests compressing the input to the encoded representation so that each input sequence can be decoded uniquely. In this case, the latent space's entropy should correspond to the input space's entropy, and compression is much more difficult.

\section{Self-Supervised Multiview Information Bottleneck Learning}
\label{sslib}

How can we learn without labels and still achieve good predictive power? Is compression necessary to obtain an optimal representation? This section analyzes and discusses how to achieve optimal representation for self-supervised learning when labels are not available during training. We review recent methods for self-supervised learning and show how they can be integrated into a single framework. We compare their objective functions, implicit assumptions, and theoretical challenges. Finally, we consider the information-theoretic properties of these representations, their optimality, and different ways of learning them.

One approach to enhance deep learning methods is to apply the \textit{InfoMax principle} in a multiview setting \citep{linsker1988self, Laurenz2002}. As one of the earliest approaches, \cite{linsker1988self} proposed maximizing information transfer from input data to its latent representation, showing its equivalence to maximizing the determinant of the output covariance under the Gaussian distribution assumption. \cite{becker1992self} introduced a representation learning approach based on maximizing an approximation of the mutual information between alternative latent vectors obtained from the same image. The most well-known application is the Independent Component Analysis (ICA) Infomax algorithm~\citep{bell1995information}, designed to separate independent sources from their linear combinations. The ICA-Infomax algorithm aims to maximize the mutual information between mixtures and source estimates while imposing statistical independence among outputs. The Deep Infomax approach~\citep{hjelm2018learning} extends this idea to unsupervised feature learning by maximizing the mutual information between input and output while matching a prior distribution for the representations. Recent work has applied this principle to a self-supervised multiview setting \citep{hjelm2018learning, henaff2020data, bachman2019learning, tian2020contrastive}, wherein these works maximize the mutual information between the views $Z_1$ and $Z_2$ using the classifier $q(z_1|z_2)$, which attempts to predict one representation from the other.

However, \cite{tschannen2019mutual} demonstrated that the effectiveness of InfoMax models is more attributable to the inductive biases introduced by the architecture and estimators than to the training objectives themselves, as the InfoMax objectives can be trivially maximized using invertible encoders. Moreover, a fundamental issue with the \textit{InfoMax principle} is that it retains irrelevant information about the labels, contradicting the core concept of the IB principle, which advocates compressing the representation to enhance generalizability.

To resolve this problem, \cite{article2008} proposed the \textit{multiview IB framework}. According to this framework, in the multiview without labels setting, the IB principle of preserving relevant data while compressing irrelevant data requires assumptions regarding the relationship between views and labels. They presented the \textit{MultiView assumption}, which asserts that either view (approximately) would be sufficient for downstream tasks. By this assumption, they define the relevant information as the shared information between the views. Therefore, augmentations (such as changing the image style) should not affect the labels.

Additionally, the views will provide most of the information in the input regarding downstream tasks. We improve generalization without affecting performance by compressing the information not shared between the two views. Their formulation is as follows:

\begin{assumption}{The \textbf{MultiView Assumption:}}
\label{ass: multiview ass}
There exists a $\epsilon_{\text{info}}$ (which is assumed to be small) such that
\begin{align*}
I(Y;X_2|X_1) &\leq \epsilon_{\text{info}},\\
I(Y;X_1|X_2) &\leq \epsilon_{\text{info}}.
\end{align*}
\end{assumption}

As a result, when the information sharing parameter, $\epsilon_\text{info}$, is small, the information shared between views includes task-relevant details. For instance, in self-supervised contrastive learning for visual data \citep{hjelm2018learning}, views represent various augmentations of the same image. In this scenario, the \textit{MultiView} assumption is considered mild if the downstream task remains unaffected by the augmentation \citep{geiping2022much}. Image augmentations can be perceived as altering an image's style without changing its content. Thus, \cite{tsai2020self} contends that the information required for downstream tasks should be preserved in the content rather than the style.
This assumption allows us to separate the information into relevant (shared information) and irrelevant (not shared) components and to compress only the unimportant details that do not contain information about downstream tasks. Based on this assumption, we aim to maximize the relevant information $I(X_2; Z_1)$ and minimize $I(X_1; Z_1 \mid X_2)$ - the exclusive information that $Z_1$ contains about $X_1$, which cannot be predicted by observing $X_2$. This irrelevant information is unnecessary for the prediction task and can be discarded. In the extreme case, where $X_1$ and $X_2$ share only label information, this approach recovers the supervised IB method without labels. Conversely, if $X_1$ and $X_2$ are identical, this method collapses into the InfoMax principle, as no information can be accurately discarded.

\cite{federici2020learning}  used the relaxed Lagrangian objective to obtain the minimal sufficient representation $Z_1$ for $X_2$ as:
\begin{align*}
    \mathcal{L}_1 = I_{\scaleto{{P(Z_1|X_1})}{6pt}}(Z_1;X_1  \mid X_2) - \beta_1 I_{\scaleto{{P(Z_2|X_2),Q(Z_1|Z_2})}{6pt}}(X_2;Z_1)
\end{align*}
and the symmetric loss to obtain the minimal sufficient representation $Z_2$ for $X_1$:
\begin{align*}
    \mathcal{L}_2 = I_{\scaleto{{P(Z_2|X_2})}{6pt}}(Z_2;X_2  \mid X_1) - \beta_2 I_{\scaleto{{P(Z_1|X_1),Q(Z_2|Z_1})}{6pt}} I(X_1;Z_2)
\end{align*}
where $\beta_1$ and $\beta_2$ are the Lagrangian multipliers introduced by the constraint optimization.
By defining $Z_1$ and $Z_2$ on the same domain and re-parameterizing the Lagrangian multipliers, the average of the two loss functions can be upper bounded as:
\begin{align*}
   \mathcal{ L} = -I_{\scaleto{{P(Z_1|X_1),Q(Z_2|Z_1})}{6pt}}(Z_1;Z_2) + \beta D_{\text{SKL}}[p(z_1  \mid x_1)||P(z_2  \mid x_2)]
\end{align*}
where $D_\text{SKL}$ represents the symmetrized $KL$ divergence obtained by averaging the expected value of $D_{\text{KL}}(p(z_1  \mid x_1)||p(z_2 \mid x_2))$ and $D_{\text{KL}}(p(z_2  \mid x_2)||p(z_1  \mid x_1))$. Note that when the mapping from $X_1$ to $Z_1$ is deterministic, $I(Z_1;X_1 \mid X_2)$ minimization and $H(Z_1  \mid X_2)$ minimization are interchangeable and the algorithms of \cite{federici2020learning} and \cite{tsai2020self} minimize the same objective. Another implementation of the same idea is based on the Conditional Entropy Bottleneck (CEB) algorithm \citep{fischer2020conditional} and proposed by \cite{lee2021compressive}. This algorithm adds the residual information as a compression term to the InfoMax objective using the reverse decoders $q(z_1  \mid  x_2)$ and $q(z_2 \mid x_1)$.

In conclusion, all the algorithms mentioned above are based on the Multiview assumption. Utilizing this assumption, they can distinguish relevant information from irrelevant information. As a result, all these algorithms aim to maximize the information (or the predictive ability) of one representation with respect to the other view while compressing the information between each representation and its corresponding view. The key differences between these algorithms lie in the decomposition and implementation of these information terms.

\cite{dubois2021lossy} offers another theoretical analysis of the IB for self-supervised learning. Their work addresses the question of the minimum bit rate required to store the input but still achieve high performance on a family of downstream tasks $Y \in \mathcal{Y}$. It is a rate-distortion problem, where the goal is to find a compressed representation that will give us a good prediction for every task. We require that the distortion measure is bounded: $$D_\mathcal{T}(X,Z) = \sup_{Y\in \mathcal{Y}} H(Y  \mid  Z_1) - H(Y  \mid  X_1) \leq \delta.$$

Accessing the downstream task is necessary to find the solution during the learning process. As a result, \cite{dubois2021lossy} considered only tasks invariant to some equivalence relation, which divides the input into disjoint equivalence classes. An example would be an image with labels that remain unchanged after augmentation. This is similar to the \textit{Multiview assumption} where $\epsilon_{info} \to 0$. By applying Shannon's rate-distortion theory, they concluded that the minimum achievable bit rate is the rate-distortion function with the above invariance distortion. Thus, the optimal rate can be determined by minimizing the following Lagrangian:

\begin{align}
\mathcal{L}= \min_{P(Z_1 \mid X_1)} I_{\scaleto{P(Z_1|X_1)}{6pt}}(X_1;Z_1)+\beta H(Z_2 \mid X_1).
\end{align}

Using this objective, the maximization of information with labels is replaced by maximizing the prediction ability of one view from the original input, regularized by direct information from the input. Similarly to the above results, we would like to find a representation $Z_1$ that compresses the input $X_1$ so that $Z_1$ has the maximum information about $X_2$.

\subsection{Implicit Compression in Self-Supervised Learning Methods}
\label{sec:compression}

While the optimal IB representation is based on the Multiview assumption, most self-supervised learning models only use the infoMax principle and maximize the mutual information $I(Z_1; Z_2)$ without an explicit regularization term. However, recent studies have shown that contrastive learning creates compressed representations that include only relevant information \citep{wang2022rethinking, tian2020makes}. The question is, why is the learned representation compressed? The maximization of $I(Z_1; Z_2)$ could theoretically be sufficient to retain all the information from both $X_1$ and $X_2$ by making the representations invertible. In this section, we attempt to explain this phenomenon.

We begin with the InfoMax principle \citep{linsker1988self}, which maximizes the mutual information between the representations of random variables $Z^1$ and $Z^2$ of the two views. We can lower-bound it using:

\begin{align}
\label{eq:lower_bound3}
I(Z_1;Z_2) = H(Z) - H(Z_1 \mid Z_2) \geq H(Z_1) + \boldsymbol{\mathbb{E}}[\log q(z_1 \mid z_2)]
\end{align}

The bound is tight when $q(z_1|z_2) = p(z_1|z_2)$, in which case the first term equals the conditional entropy $H(Z_1|Z_2)$. The second term of \cref{eq:lower_bound3} can be considered a negative reconstruction error or distortion between $Z_1$ and $Z_2$.

In the supervised case, where $Z$ is a learned stochastic representation of the input and $Y$ is the label, we aim to optimize

\begin{equation}
I(Y; Z) \geq H(Y) + \boldsymbol{\mathbb{E}}\left[\log q(Y\mid Z)\right]
\end{equation}
. Since $Y$ is constant, optimizing the information $I(Z; Y)$ requires only minimizing the prediction term $\mathbb{E}\left[\log q(Y|Z)\right]$ by making $Z$ more informative about $Y$. This term is the cross-entropy loss for classification or the square loss for regressions. Thus, we can minimize the log loss without any other regularization on the representation.

In contrast, for the self-supervised case, we have a more straightforward option to minimize $H(Z_1|Z_2)$: Making $Z_1$ easier to predict by $Z_2$, which can be achieved by reducing its variance along specific dimensions. If we do not regularize $H(Z_1)$, it will decrease to zero, and we will observe a collapse. This is why, in contrastive methods, the variance of the representation (large entropy) is significant only in the directions with a high variance in the data, which is enforced by data augmentation \citep{jing2021understanding}. According to this analysis, the network benefits from making the representations "simple" (easier to predict). Hence, even though our representation does not have explicit information-theoretical constraints, the learning process will compress the representation.

\subsection{Beyond the Multiview Assumption}
\label{sec:beyond_multiview_assumption}

According to the Multiview IB analysis presented in Section \ref{sslib}, the optimal way to create a useful representation is to maximize the mutual information between the representations of different views while compressing irrelevant information in each representation. In fact, as discussed in Section \ref{sec:compression}, we can achieve this optimal compressed representation even without explicit regularization. However, this optimality is based on the \textit{Multiview assumption}, which states that the relevant information for downstream tasks comes from the information shared between views. Therefore, \cite{tian2020makes} concluded that when a minimal sufficient representation has been obtained, the optimal views for self-supervised learning are determined by downstream tasks.

However, the \textit{Multiview assumption} is highly constrained, as all relevant information must be shared between all views. In cases where this assumption is incorrect, such as with aggressive data augmentation or multiple downstream tasks or modalities, sharing all the necessary information can be challenging. For example, if one view is a video stream while the other is an audio stream, the shared information may be sufficient for object recognition but not for tracking. Furthermore, relevant information for downstream tasks may not be contained within the shared information between views, meaning that removing non-shared information can negatively impact performance.

\cite{kahana2022contrastive} identified a series of tasks that violate the \textit{Multiview assumption}. To accomplish these tasks, the learned representation must also be invariant to unwanted attributes, such as bias removal and cross-domain retrieval. In such cases, only some attributes have labels, and the objective is to learn an invariant representation for the domain for which labels are provided while also being informative for all other attributes without labels. For example, for face images, only the identity labels may be provided, and the goal is to learn a representation that captures the unlabeled pose attribute but contains no information about the identity attribute. The task can also be applied to fair decisions, cross-domain matching, model anonymization, and image translation.

\cite{wang2022rethinking} formalized another case where the \textit{Multiview assumption} does not hold when non-shared task-relevant information cannot be ignored. In such cases, the minimal sufficient representation contains less task-relevant information than other sufficient representations, resulting in inferior performance. Furthermore, their analysis shows that in such cases, the learned representation in contrastive learning is insufficient for downstream tasks, which may overfit the shared information.

As a result of their analysis, \cite{wang2022rethinking} and \cite{kahana2022contrastive} proposed explicitly increasing mutual information between the representation and input to preserve task-relevant information and prevent the compression of unshared information between views. In this case, the two regularization terms of the two views are incorporated into the original InfoMax objective, and the following objective is optimized:

\begin{align}
\label{eq:reconst}
\mathcal{L} = \min_{P(Z_1|X_1), p(Z_2|X_2)} -I_{\scaleto{P(Z_1|X_1)}{6pt}}(X_1;Z_1) -I_{\scaleto{P(Z_2|X_2)}{6pt}}(X_2;Z_2) -\beta I_{\scaleto{P(Z_1|X_1),P(Z_2|Z_1)}{6pt}}(Z_1;Z_2) .
\end{align}
\cite{wang2022rethinking} demonstrated the effectiveness of their method for SimCLR \citep{chen2020simple}, BYOL \citep{grill2020bootstrap}, and Barlow Twins \citep{zbontar2021barlow} across classification, detection, and segmentation tasks.
\subsection{To Compress or Not to Compress?}
As seen in Eq. \ref{eq:reconst}, when the \textit{Multiview assumption} is violated, the objective for obtaining an optimal representation is to \textbf{maximize} the mutual information between each input and its representation. This contrasts with the situation in which the \textit{Multiview assumption} holds, or the supervised case, where the objective is to \textbf{minimize} the mutual information between the representation and the input. In both supervised and unsupervised cases, we have direct access to the relevant information, which we can use to separate and compress irrelevant information. However, in the self-supervised case, we depend heavily on the \textit{Multiview assumption}. If this assumption is violated due to unshared information between views that is relevant for the downstream task, we cannot separate relevant and irrelevant information. Furthermore, the learning algorithm's nature requires that this information be protected by explicitly maximizing it.

As datasets continue to expand in size and models are anticipated to serve as base models for various downstream tasks, the \textit{Multiview assumption} becomes less pertinent. Consequently, compressing irrelevant information when the \textit{Multiview assumption} does not hold presents one of the most significant challenges in self-supervised learning. Identifying new methods to separate relevant from irrelevant information based on alternative assumptions is a promising avenue for research. It is also essential to recognize that empirical measurement of information-theoretic quantities and their estimators plays a crucial role in developing and evaluating such methods.

\section{Optimizing Information in Deep Neural Networks: Challenges and Approaches}

\label{sec:estimators}

Recent years have seen information-theoretic analyses employed to explain and optimize deep learning techniques \citep{shwartz2017opening}. Despite their elegance and plausibility, empirically measuring and analyzing information in deep networks presents challenges. Two critical problems are (1) information in deterministic networks and (2) estimating information in high-dimensional spaces.

\subsubsection*{\textbf{Information in Deterministic Networks}}

\label{section:inf_deterministic}

Information-theoretic methods have significantly impacted deep learning \citep{vib, steinke2020reasoning, shwartz2017opening}. However, a key challenge is addressing the source of randomness in deterministic DNNs.

The mutual information between the input and representation is infinite, leading to ill-posed optimization problems or piecewise constant outcomes \citep{amjad2019learning, 2018Estimating}. To tackle this issue, researchers have proposed various solutions. One common approach is to discretize the input distribution and real-valued hidden representations by binning, which facilitates non-trivial measurements and prevents the mutual information from always taking the maximum value of the log of the dataset size, thus avoiding ill-posed optimization problems \citep{shwartz2017opening}.

However, binning and discretization are essentially equivalent to geometrical compression and serve as clustering measures \citep{2018Estimating}. Moreover, this discretization depends on the chosen bin size and does not track the mutual information across varying bin sizes \cite{2018Estimating, ross2014mutual}. To address these limitations, researchers have proposed alternative approaches such as interpreting binned information as a weight decay penalty \cite{elad2019the}, estimating mutual information based on lower bounds assuming a continuous input distribution without making assumptions about the network's output distribution properties \citep{wang2020understanding, zimmermann2021contrastive, shwartz2022we}, injecting additive noise, and considering data augmentation as the source of noise \citep{lee2021compressive, shwartz2017opening, 2018Estimating, dubois2021lossy}.

\subsubsection*{\textbf{Measuring Information in High-Dimensional Spaces}}

\label{section:inf_determnistic}

Estimating mutual information in high-dimensional spaces presents a significant challenge when applying information-theoretic measures to real-world data. This problem has been extensively studied \citep{Paninski:2003:EEM:795523.795524, gao2015efficient}, revealing the inefficiency of solutions for large dimensions and the limited scalability of known approximations with respect to sample size and dimension. Despite these difficulties, various entropy and mutual information estimation approaches have been developed, including classic methods like k-nearest neighbors (KNN) \citep{kozachenko1987sample} and kernel density estimation techniques \citep{hang2018kernel}, as well as more recent efficient methods. 

\cite{chelombiev2019adaptive} developed adaptive mutual information estimators based on entropies-equal bins and scaled noise kernel density estimator. Generative decoder networks, such as PixelCNN++ \citep{van2016conditional}, have been employed to estimate a lower bound on mutual information \citep{darlow2020information, nash2018inverting, shwartz2023information}. Another strategy includes ensemble dependency graph estimators, adaptive mutual information estimation methods (EDGE) by merging randomized locality-sensitive hashing (LSH), dependency graphs, and ensemble bias reduction techniques \citep{noshad2018scalable}. The Mutual Information Neural Estimator (MINE) \citep{Belghazi2018MutualIN} maximizes KL divergence using the dual representation of \cite{donsker1975asymptotic} and has been employed for direct mutual information estimation \citep{elad2019direct}. \cite{shwartz2020information} developed a controlled framework that utilized the neural tangent kernels \citep{jacot2018neural},  in order to obtain tractable information measures.

Improving mutual information estimation can be achieved using larger batch sizes, although this may negatively impact generalization performance and memory requirements. Alternatively, researchers have suggested employing surrogate measures for mutual information, such as log-determinant mutual information (LDMI), based on second-order statistics \citep{ozsoy2022self, erdogan2022information}, which reflects linear dependence. \cite{goldfeld2021sliced} proposed the Sliced Mutual Information (SMI), defined as an average of MI terms between one-dimensional projections of high-dimensional variables. SMI inherits many properties of its classic counterpart. It can be estimated with optimal parametric error rates in all dimensions by combining an MI estimator between scalar variables with an MC integrator \citep{goldfeld2021sliced}. The $k$-SMI, introduced by \cite{goldfeld2022k}, extends the SMI by projecting to $k$-dimensional subspace, which relaxes the smoothness assumptions, improves scalability, and enhances performance.

In conclusion, estimating and optimizing information in deep neural networks presents significant challenges, particularly in deterministic networks and high-dimensional spaces. Researchers have proposed various approaches to address these issues, including discretization, alternative estimators, and surrogate measures. As the field continues to evolve, it is expected that more advanced techniques will emerge to overcome these challenges and facilitate the understanding and optimization of deep learning models.
\section{Future Research Directions}
\label{sec:future_research}

Despite the solid foundation established by existing self-supervised learning methods from an information theory perspective, several potential research directions warrant exploration:

\paragraph{Self-supervised learning with non-shared information.} As discussed in Section \ref{sslib}, the separation of relevant (preserved) and irrelevant (compressed) information relies on the \textit{Multiview Assumption}. This assumption, which states that only shared information is essential for downstream tasks, is rather restrictive. For example, situations may arise where each view contains distinct information relevant to a downstream task or multiple tasks necessitate different features. Some methods have been proposed to tackle this problem, but they mainly focus on maximizing the network's information without explicit constraints. Formalizing this scenario and exploring differentiating between relevant and irrelevant data based on non-shared information represents an intriguing research direction.

\paragraph{Self-supervised learning for tabular data.} At present, the internal compression of self-supervised learning methods may compress relevant information due to improper augmentation \ref{sec:compression}. Consequently, we must heavily rely on generating the two views, which must accurately represent information related to the downstream process. Custom augmentation must be developed for each domain, taking into account extensive prior knowledge on data augmentation. While some papers have attempted to extend self-supervised learning to tabular data \citep{ucar2021subtab, arik2021tabnet}, further work is necessary from both theoretical and practical standpoints to achieve high performance with self-supervised learning for tabular data \citep{shwartz2022tabular}. The augmentation process is crucial for the performance of current vision and text models. In the case of tabular data, employing information-theoretic loss functions that do not require information compression may help harness the benefits of self-supervised learning.

\paragraph{Integrating other learning methods into the information-theoretic framework.} Prior works have investigated various supervised, unsupervised, semi-supervised, and self-supervised learning methods, demonstrating that they optimize information-theoretic quantities. However, state-of-the-art methods employ additional changes and engineering practices that may be related to information theory, such as the stop gradient operation utilized by many self-supervised learning methods today \citep{grill2020bootstrap, chen2021exploring}. The Expectation-Maximization (EM) algorithm \citep{dempster1977maximum} can be employed to explain this operation when one path is the E-step and the other is the M-step. Additionally, \cite{elidan2012information} proposed an IB-inspired version of the EM, which could help develop information-theoretic-based objectives using the stop gradient operation.

\paragraph{Expanding the analysis to usable information.} While information theory offers a rigorous conceptual framework for describing information, it neglects essential aspects of computation. (Conditional) entropy, for example, is directly related to the predictability of a random variable in a betting game where agents are rewarded for accurate guesses. However, the standard definition assumes that agents have no computational bounds and can employ arbitrarily complex prediction schemes \citep{cover1999elements}. In the context of deep learning, predictive information $H(Y|Z)$ measures the amount of information that can be extracted from $Z$ about $Y$ given access to all decoders $p(y|z)$ in the world. Recently, \cite{xu2020theory} introduced \textit{predictive V-information} as an alternative formulation based on realistic computational constraints.

\paragraph{Extending self-supervised learning's information-based perspective to energy-based model optimization.} Until now, research combining self-supervised learning with information theory has focused on probabilistic models with tractable likelihoods. These models enable specific optimization of model parameters concerning the tractable log-likelihood \citep{graves2013generating, germain2015made, dinh2016density, rezende2015variational} or a tractable lower bound of the likelihood \citep{kingma2019introduction, vib}. Although models with tractable likelihoods offer certain benefits, their scope is limited and necessitates a particular format. Energy-based models (EBMs) present a more flexible, unified framework. Rather than specifying a normalized probability, EBMs define inference as minimizing an unnormalized energy function and learning as minimizing a loss function. The energy function does not require integration and can be parameterized with any nonlinear regression function. Inference typically involves finding a low-energy configuration or sampling from all possible configurations such that the probability of selecting a specific configuration follows a Gibbs distribution \citep{huembeli2022physics, song2021train}.

Investigating energy-based models for self-supervised learning from both theoretical and practical perspectives can open up numerous promising research directions. For instance, we could directly apply tools developed for energy-based models and statistical machines to optimize the model, such as Maximum Likelihood Training with MCMC \citep{Younes99onthe}, score matching \citep{Hyvrinen06someextensions}, denoising score matching \citep{song2020score, Vincent2011}, and score-based generation models \citep{NEURIPS2019_3001ef25}.

\paragraph{Expanding the multiview framework to accommodate more views and tasks.} The multiview self-supervised IB framework can be extended to cases involving more than two views $(X_1, \cdots, X_n)$ and multiple downstream tasks $(Y_1, \cdots, Y_K)$. A simple extension of the multiview IB framework can be achieved by setting the objective function to maximize the joint mutual information of all views' representations $I(Z_1; \cdots Z_n)$ and compressing the individual information for each view $I(X_i; Z_i), \quad 1 \leq i \leq N
$ However, to ensure the optimality of this objective, we must expand the \textit{multiview assumption} to include more than two views. In this scenario, we need to assume that relevant information is shared among all different views and tasks, which might be overly restrictive. As a result, defining and analyzing a more refined version of this naive solution is essential. One potential approach involves utilizing the Multi-feature Information Bottleneck (MfIB) \citep{lou2013multi}, which extends the original IB. The MfIB processes multiple feature types simultaneously and analyzes data from various sources. This framework establishes a joint distribution between the multivariate data and the model. Rather than solely preserving the information of one feature variable maximally, the MfIB concurrently maintains multiple feature variables' information while compressing them. The MfIB characterizes the relationships between different sources and outputs by employing the multivariate Information Bottleneck \citep{friedman2013multivariate} and setting Bayesian networks.
\section{Conclusion}
\label{sec:conclusion}

In this study, we delved deeply into the concept of optimal representation in self-supervised learning through the lens of information theory. We synthesized various approaches, highlighting their foundational assumptions and constraints, and integrated them into a unified framework. Additionally, we explored the key information-theoretic terms that influence these optimal representations and the methods for estimating them.

While supervised and unsupervised learning offer more direct access to relevant information, self-supervised learning depends heavily on assumptions about the relationship between data and downstream tasks. This reliance makes distinguishing between relevant and irrelevant information considerably more challenging, necessitating further assumptions.

Despite these challenges, information theory stands out as a robust and versatile framework for analysis and algorithmic development. This adaptable framework caters to a range of learning paradigms and elucidates the inherent assumptions underpinning data and model optimization.

With the rapid growth of datasets and the increasing expectations placed on models to handle multiple downstream tasks, the traditional Multi-view assumption might become less reliable. One significant challenge in self-supervised learning is the precise compression of irrelevant information, especially when these assumptions are compromised.

Future research avenues might involve expanding the Multi-view framework to include more views and tasks and deepening our understanding of information theory's impact on facets of deep learning, such as reinforcement learning and generative models.

In summary, information theory is a crucial tool in our quest to understand better and optimize self-supervised learning models. By harnessing its principles, we can more adeptly navigate the intricacies of deep neural network development, paving the way for creating more effective models.

-

%% CONCLUSION
%\input{drafts/_arxiv/4_conclusion}

%%%%%%%%%%%%%%%%%%%%%%%%%%%%%%

%% ACKNOWLEDGEMENTS
%\input{drafts/_arxiv/5_acknowledgements}

\clearpage

%% REFERENCES
\bibliography{__main}

\clearpage

%% APPENDICES
%\input{drafts/_arxiv/6_appendices}

\end{document}